\documentclass[conference]{IEEEtran}
\IEEEoverridecommandlockouts
\usepackage{cite}
\usepackage{amsmath,amssymb,amsfonts}
\usepackage{algorithmic}
\usepackage{graphicx}
\usepackage{textcomp}
\usepackage{xcolor}

\usepackage{algorithm}
\usepackage{algorithmic}
\usepackage{multirow}
\usepackage{amsmath,amssymb,amsfonts}
\usepackage{booktabs}
\usepackage{multirow}
\usepackage{subfigure} 
\usepackage{newfloat}
\usepackage{listings}

\def\BibTeX{{\rm B\kern-.05em{\sc i\kern-.025em b}\kern-.08em
    T\kern-.1667em\lower.7ex\hbox{E}\kern-.125emX}}
\begin{document}

\title{Cross-Task Attack: A Self-Supervision Generative Framework Based on  Attention Shift\\
}

\author{\IEEEauthorblockN{1\textsuperscript{st} Qingyuan Zeng}
\IEEEauthorblockA{\textit{Institute of Artificial Intelligence} \\
\textit{Xiamen University}\\
Fujian, China \\
36920221153145@stu.xmu.edu.cn}
\and
\IEEEauthorblockN{2\textsuperscript{nd} Yunpeng Gong}
\IEEEauthorblockA{\textit{School of Informatics} \\
\textit{Xiamen University}\\
Fujian, China \\
fmonkey625@gmail.com}
\and
\IEEEauthorblockN{3\textsuperscript{rd} Min Jiang*}
\IEEEauthorblockA{\textit{School of Informatics} \\
\textit{Xiamen University}\\
Fujian, China \\
minjiang@xmu.edu.cn}
\thanks{The corresponding author: Min Jiang, minjiang@xmu.edu.cn}
}

\maketitle

\begin{abstract}
Studying adversarial attacks on artificial intelligence (AI) systems helps discover model shortcomings, enabling the construction of a more robust system. Most existing adversarial attack methods only concentrate on single-task single-model or single-task cross-model scenarios, overlooking the multi-task characteristic of artificial intelligence systems.   As a result, most of the existing attacks do not pose  a practical threat to a comprehensive  and collaborative AI system. However, implementing cross-task attacks is highly demanding and challenging due to the difficulty in obtaining the real labels of different tasks for the same picture and harmonizing the loss functions across different tasks. To address this issue, we propose a self-supervised Cross-Task Attack framework (CTA), which utilizes co-attention and anti-attention maps to generate cross-task adversarial perturbation. Specifically, the co-attention map reflects the area to which different visual task models pay attention, while the anti-attention map reflects the area that different visual task models neglect. CTA generates cross-task perturbations by shifting the attention area of samples away from the co-attention map and closer to the anti-attention map.  We conduct extensive experiments on multiple vision tasks and the experimental results confirm the effectiveness of the proposed design for adversarial attacks.
\end{abstract}


\section{Introduction}

In recent years, the widespread application of artificial intelligence (AI) systems has brought dramatic changes in various fields. As AI technologies become more pervasive in our daily lives, people start to worry about their robustness and safety. Adversarial attacks \cite{Goodfellow2014, Kurakin2016, Dezfooli2016,Gong_2021_arXiv,Gong_2022_CVPRW,Gong_2024_arXiv,Gong_2024_arXiv2,jiang2017integration,jiang2021individual,jiang2021knee,wang2024evolutionary,jiang2017transfer} are techniques that use small perturbations imperceptible to humans to deceive AI systems, and have become an important research topic. The goal is to reveal model weaknesses and help developers build more robust systems. To provide a strong baseline for deep learning robustness research, many studies have proposed various effective attack methods to generate adversarial samples.

\begin{figure}[t]
\centering
\includegraphics[width=0.7\columnwidth]{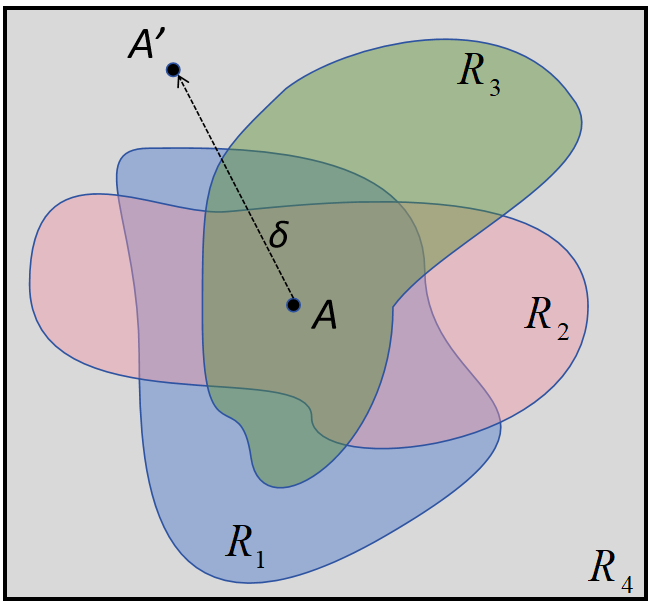}
\caption{Schematic illustration of the proposed idea for cross-task attacks. In the schematic diagram, $R_1$, $R_2$ and $R_3$ represent the attention regions of the input image for three different visual tasks, respectively, and $R_4$ represents all regions of the image. Co-attention represents the union of $R_1$, $R_2$, and $R_3$, while anti-attention represents the complement of co-attention in $R_4$. $A$ represents the attention point of the original sample, located within the co-attention area, so the original image can be accurately recognized by all visual tasks. Conversely, $A'$ represents the attention point of the adversarial sample, located within the anti-attention area, so the adversarial sample can effectively evade recognition  of all visual tasks.
}
\label{fig:r1r2r3}
\end{figure}

Existing adversarial attack methods can be classified according to three criteria: 1. sample-specific or cross-sample, 2. model-specific or cross-model, 3. task-specific or cross-task. Extensive research has been conducted on the first two criteria.  Some researchers conducted pioneering research on cross-sample attacks, aiming to obtain a perturbation that can disturb multiple samples simultaneously \cite{Moosavi-Dezfooli2017, Mopuri2017, Mopuri2018, Zhang_2021_ICCV, Peng_2022_CVPR}. On the other hand,  some researchers conducted research on cross-model attacks, aiming to improve the transferability  of adversarial perturbations by adding additional randomness  \cite{Xie2018, Dong2019, Wang2021, Wang2021a,Zhang2021, Long2022}. Cross-sample attacks can speed up the production of adversarial example sets, while cross-model attacks can increase the possibility of black-box attacks under specific tasks. 

Most existing research on adversarial attacks mainly focuses on single-task scenarios, ignoring the multi-task characteristics of AI systems. In practical applications, AI systems need to cooperate with multiple tasks for decision-making \cite{Kim2020, Lee2021, Feng2021}. Neither cross-sample nor cross-model attacks can effectively threaten AI systems in practical applications. 

Unlike cross-sample and cross-model attacks, the core of the cross-task adversarial attack methods is to find and destroy the common characteristics of different vision tasks.  DR \cite{Lu2020} first proposed a cross-task attack method called Dispersion Reduction. DR considers that the common characteristics of different vision tasks is the feature extractor. However, the attack performance of DR is weak because the feature extractors vary greatly among different tasks and models.

\begin{figure*}[t]
\centering
\includegraphics[width=2\columnwidth]{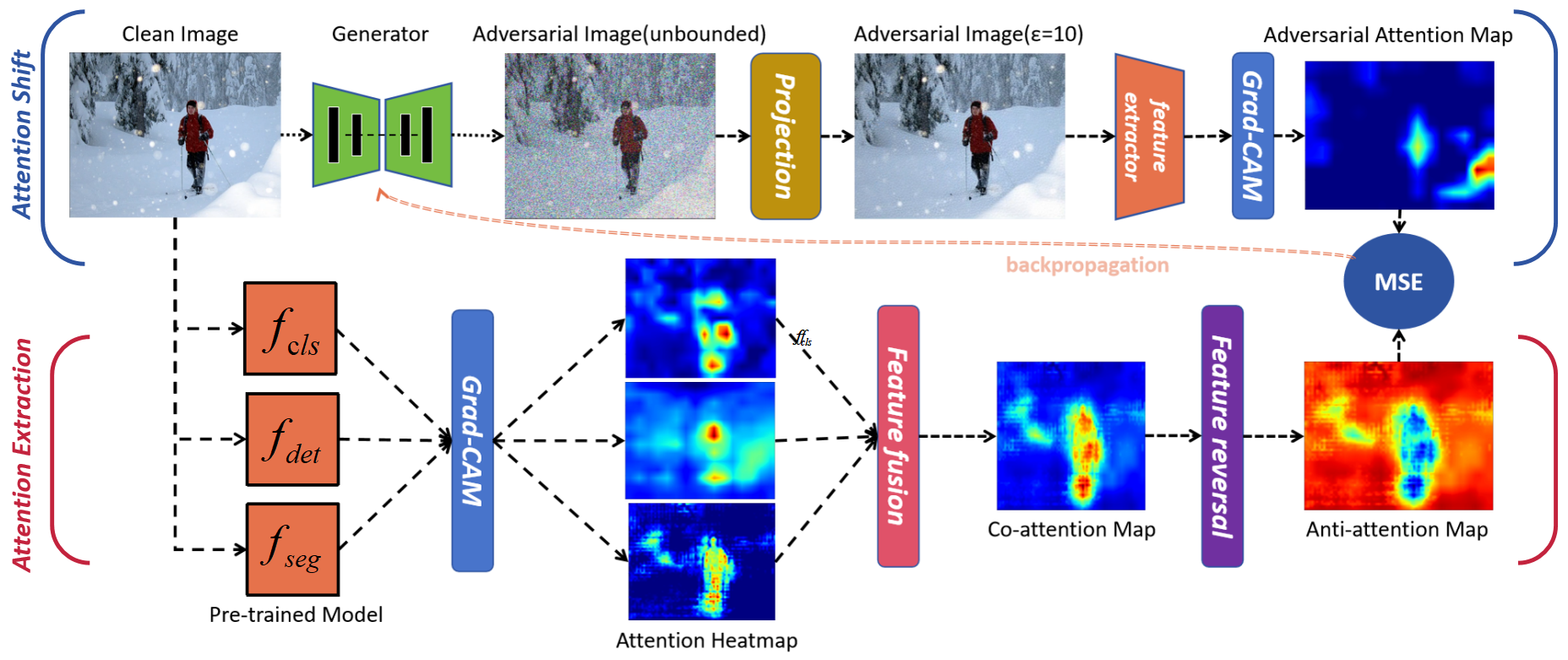}
\caption{The framework diagram about our proposed self-supervised cross-task attack method. We use existing pre-trained models to extract the anti-attention map of the input image as the ground-true label of the framework. We use the generator to generate adversarial perturbation to change the mapping of image in feature space. By shortening the MSE distance between the adversarial attention map and the anti-attention map, the attention area of the adversarial image falls in the area that is ignored by all visual tasks.}
\label{fig:framework}
\end{figure*}

In this paper, we propose a self-supervision generative framework based on attention shift to enable cross-task attack. Our approach, called Cross-Task Attack (CTA) and illustrated in Figure \ref{fig:framework}, is inspired by previous explorations around the principles of adversarial attacks \cite{Wang2021b, Chakraborty2022}.  Adversarial samples can fool neural networks because the perturbations make the models’ attention move to unimportant areas \cite{Chakraborty2022}. So we presume that cross-task attack can be achieved by directing the attention of the models to areas that all visual tasks neglect. 

We use co-attention map to represent the regions that multiple visual tasks focus on, while anti-attention map to represent the regions  that all visual tasks neglect. As shown in Figure \ref{fig:r1r2r3}, co-attention is the union  of attention regions from different visual tasks, while anti-attention is the complement of co-attention.  By using perturbation $\delta$ to shift the attention of adversarial sample from point $A$ in co-attention region to point $A'$ in anti-attention region, cross-task attack can be achieved. It is worth noting that co-attention and anti-attention maps are obtained using ready-made pre-trained models, so CTA does not require any ground truth labels for training.

\begin{figure}[t]
\centering
\includegraphics[width=1\columnwidth]{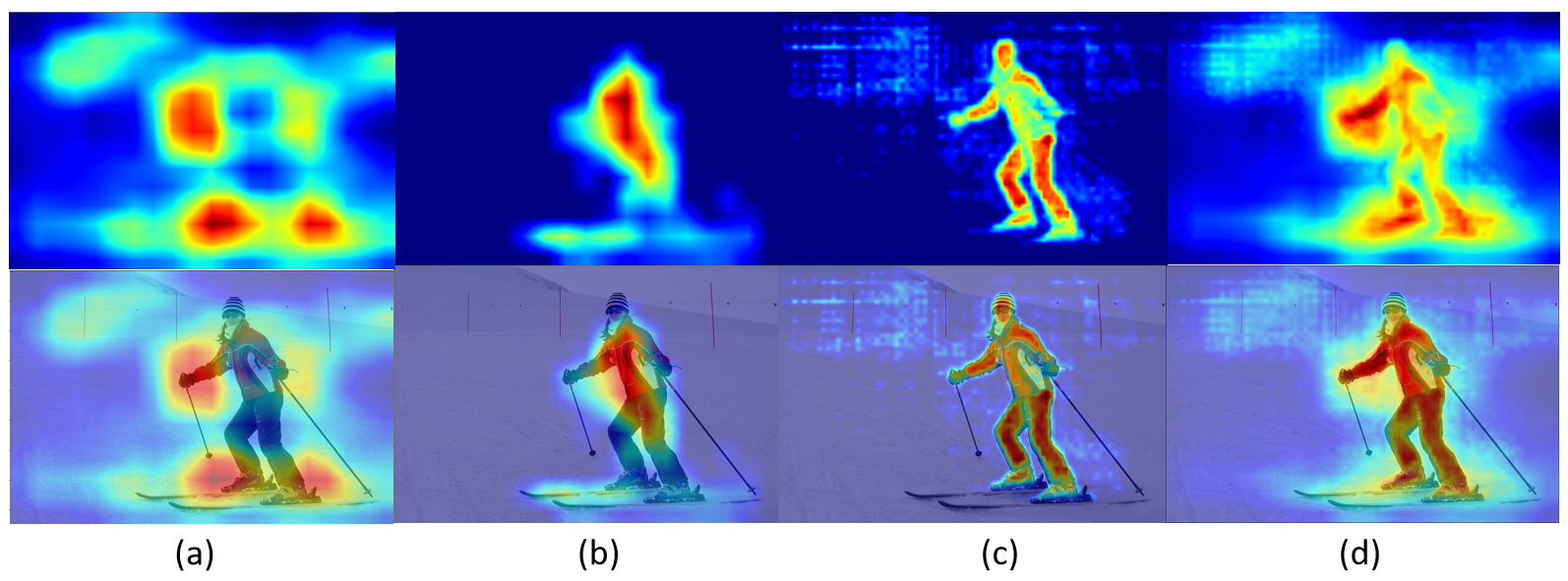}
\caption{Grad-CAM heatmaps for different visual tasks are displayed. The first row presents the heatmaps, while the second row overlays these heatmaps onto the original image. The column (a) is the classification task based on ResNet50, the column (b) is the semantic segmentation task based on DeepLabv3, the column (c) is the object detection task based on Faster-RCNN, and the 
 column (d) is the co-attention heatmap  that all visual tasks focus on.}
\label{fig:tasks_gradcam}
\end{figure}

Based on the experimental conclusions of previous work \cite{Wang2021b}, attention heatmap is a model-agnostic shared feature in specific task. As shown in Figure \ref{fig:tasks_gradcam}, we can see that the attention heatmaps of different tasks are very different, which means that attention heatmap is shared in a specific task, but not shared in different tasks. This explains why adversarial examples based on single-task attacks fail on other tasks, because single-task attacks only divert the attention of adversarial examples from the attention area of a specific task, but may be moved to attention area of other tasks.

\textbf{Contributions.} The main contributions of this paper are as follows: 

\begin{itemize}
\item We conduct an intuitive principle analysis of existing single-task and cross-task attack methods, explaining their weaker performance in cross-task scenarios from the perspective of attention.

\item We are the first to apply common attention from different visual tasks in adversarial attacks. We propose a self-supervised generative framework CTA to shift the attention of images to regions that are overlooked by variable visual task models, enabling cross-task attack.

\end{itemize}

\section{Related Works}

\subsection{Single-task Single-model Attack}
Single-task single-model attack means that the attacker designs an adversarial example that can deceive the target model while knowing the parameters and structure of the target model. The basic idea of single-task single-model attack can be divided into two categories, one is to use gradient ascent  in the image pixel space to maximize the loss function \cite{Goodfellow2014, Fawzi2015, Fawzi2016, Kurakin2016}, and the other is to use complex optimization process to find the optimal solution leading to wrong prediction  \cite{Dezfooli2016, su2019one}. 

\subsection{Single-task Cross-model Attack}

In order to enable adversarial samples to attack different  models under specific tasks, many studies have investigated how to improve the transferability of adversarial samples.  DIM \cite{Xie2018} incorporated random transformations in the gradient iteration process to increase  the diversity of adversarial perturbations. TI-FGSM \cite{Dong2019} added a translational data augmentation method to increase the translational invariance of the adversarial examples.  SI-FGSM \cite{lin2019nesterov} utilized the scalability invariance of deep learning models and proposes adding random scaling in the gradient iteration process. DAS \cite{Wang2021b} achieved single-task attack by suppressing Grad-CAM heatmaps \cite{Selvaraju2017}. $\text{S}^{2}$I-FGSM \cite{Long2022} applies spectral transformation in the frequency domain to enhance the mobility of adversarial samples.   These works all rely on the loss function of a specific task and they cannot guide the movement direction of the attention area of the adversarial example, which makes them unable to attack other visual tasks.

\subsection{Cross-task Attack}
DR \cite{Lu2020} introduced a method for generating adversarial examples without relying on specific-task loss functions. By utilizing VGG \cite{simonyan2014very} to extract image feature maps and reducing the standard deviation, they obtained adversarial examples that can disrupt feature extraction, thus achieving cross-task attacks.  RB \cite{Zhang2021} proposed random blur (RB) during iterative optimization against perturbations, which improves the diversity of adversarial perturbations. RB can slightly improve the performance of DR in scenarios of cross-task attacks. These works are cross-task attack methods that do not depend on task-specific loss functions, but their cross-task attack performance is weak because they cannot guide the attention shift direction of adversarial samples.  

Different from the idea of perturbing feature extractors in the existing cross-task attack methods, we solve the problem of cross-task attack from a novel perspective. We pioneered the concepts of co-attention and anti-attention maps, and utilized them to guide the direction of attention-shift for adversarial samples. The attention of the adversarial examples is shifted to regions that are not concerned by various vision tasks to enable cross-task attacks.

\section{The proposed method}

\begin{algorithm}[tb]
\caption{Cross-Task Attack Method}
\label{alg:CTA}
\textbf{Input}: Clean image $x$, classification model $f_{cls}$,  detection model $f_{det}$, segmentation model $f_{seg}$, generator $G$, feature extractor $D$
\begin{algorithmic}[1] 
\STATE Initialize generator $G$ with random weights.
\FOR{$i$=1 to $T$}
\STATE Calculate the Grad-CAM maps $\mathcal{A}_c$, $\mathcal{A}_d$, and $\mathcal{A}_s$ of $f_{cls}$, $f_{det}$, and $f_{seg}$ for clean samples through Eq.\eqref{eq:eq4}.
\STATE The co-attention map is obtained by merging the features of $\mathcal{A}_c$, $\mathcal{A}_d$, and $\mathcal{A}_s$ through Eq.\eqref{eq:eq5}.
\STATE The anti-attention map is obtained by inverting the co-attention through Eq. \eqref{eq:eq6}.
\STATE Use generator $G$ to change the mapping of input image $x$ in the feature space and obtain the unbounded adversarial image $x'$. The calculation process is as in Eq.\eqref{eq:eq7}.
\STATE Restrict the distribution of $x'$ to the range of [$x-\epsilon$, $x+\epsilon$] and obtain the bounded adversarial image $x_{adv}$ through Eq.\eqref{eq:eq8}.
\STATE Use the pre-trained feature extractor $D$ to extract the features of $x_{adv}$ and obtain the category $y^c$ with the highest confidence. Substitute Eq.\eqref{eq:eq4} to calculate the $\mathcal{A}_{adv}$ of $x_{adv}$. 
\STATE Update parameters of $G$ using Adam optimizer to minimize the loss function Eq.\eqref{eq:eq9}.
\ENDFOR
\STATE Return generator $G$.
\end{algorithmic}
\end{algorithm}

In this section, we introduce how our proposed Cross-Task Attack (CTA) 
 shifts the attention of adversarial samples from important areas to areas that are overlooked by various visual tasks.

\subsection{Overview of the Framework}
In order to obtain cross-task adversarial samples, we first need to identify the regions of the samples that are not of interest to various visual tasks, and then use adversarial perturbations to shift attention to these regions. We propose a self-supervised cross-task attack method named CTA, as shown in Figure \ref{fig:framework}. CTA consists of two stages: attention extraction stage and attention shift stage. The attention extraction stage is to obtain  the co-attention and anti-attention maps of clean samples. The former reflects the important areas that different vision task models need to pay attention to, while the latter reflects the unimportant areas that are ignored. The attention shift stage is to shift the adversarial samples' attention from the co-attention area to the anti-attention area by adding adversarial perturbations, thus enabling cross-task attacks.

\subsection{Attention Extraction Stage}
The process of attention extraction stage is the bottom part of Figure \ref{fig:framework}. First, we use ready-made pre-trained models and Grad-CAM to obtain attention maps of clean samples in different vision tasks. Because the attention of different models of the same task is similar, it is enough to choose one model for each vision task \cite{Wang2021b}.  Specifically, the formula for calculating the Grad-CAM attention heatmap $\mathcal{A}$ is 

\begin{equation}
    \mathcal{A}(i, j)=\operatorname{max}\left(0,  \frac{1}{Z}\sum_{k}   \sum_{i} \sum_{j}  \cdot \frac{\partial y^{c}}{\partial \mathcal{F}_{k}(i,j)} \cdot \mathcal{F}_{k}(i,j)\right),
\label{eq:eq4}
\end{equation}
where $Z$ is the total number of pixels in the feature map, $y^{c}$ is the probability that classifier $f_{cls}$ predicts that the input image $x$ belongs to class $c$, and $\mathcal{F}_{k}(i,j)$ is the value of the $k$-th feature map of the last convolutional layer at position $(i,j)$. 

After obtaining the attention map $\mathcal{A}$ for each vision task, we need to find which regions of the picture are the focus of all vision tasks. Therefore, we fuse the attention heatmaps of different visual tasks to obtain the co-attention map, which represents the common focus area of different visual tasks. We used a simple and effective method for feature fusion, calculated as follows:

\begin{equation} 
   \text{co-attention}(i, j) =  \operatorname{Scale}\left(\frac{1}{K} \sum_{k}   \mathcal{A}_{k}(i,j)\right),
\label{eq:eq5}
\end{equation}
where $Scale$ means to normalize the value range of the heatmap to [0,1], $K$ represents the number of visual tasks, and $\mathcal{A}_k(i,j)$ represents the value of attention heatmap of the $k$-th visual task at position $(i,j)$. The high-value pixel area of co-attention map is the area that different visual tasks all focus on.

At last, we invert the co-attention map to get the anti-attention map as follow: 

\begin{equation}
    \text{anti-attention} (i, j) = 1 - \text { co-attention }(i, j),
\label{eq:eq6}
\end{equation}
anti-attention represents regions that are not attended to by all vision tasks, which can be used as labels for self-supervised training in  attention shift stage. 

\subsection{Attention Shift Stage}
The process of attention shift stage is the upper part of Figure \ref{fig:framework}.  To shift the attention of  input image, we use a generator to generate adversarial perturbations and add them to the image to change its mapping in the feature space. The calculation process for adversarial sample is as follows:

\begin{equation}
x' = x + G(x),
\label{eq:eq7}
\end{equation}
where $x$ represents the input clean sample, $x'$ represents the adversarial sample without range constraints,  $G$ represents the generator. To increase the invisibility of adversarial samples, we need to crop the adversarial sample at the pixel level:

\begin{equation}
x_{adv}(i,j) = \operatorname{min}(x(i,j)+\epsilon , \operatorname{max}(x'(i,j), x(i,j)-\epsilon )),
\label{eq:eq8}
\end{equation}
where $x_{adv}(i,j)$ and $x(i,j)$ represents the value of adversarial sample and clean sample at position $(i,j)$, $\epsilon$ represents disturbance range threshold. Each pixel of the adversarial sample $x_{adv}$ is cropped to the range of [$x-\epsilon$, $x+\epsilon$].

In order to obtain the attention heatmaps of adversarial samples, we used the parameter-frozen feature extractor (ResNet50) to calculate $y^c$. By substituting $y^c$ into Equation 4, the attention  map $\mathcal{A}_{adv}$ of the adversarial sample can be obtained. We use the distance between anti-attention map and  $\mathcal{A}_{adv}$ as the loss function to update the parameters of generator $G$. More precisely, the loss function is

\begin{equation}
\mathcal{L} = \frac{1}{N}  \sum_{i,j}(\text{anti-attention}(i,j) - \mathcal{A}_{adv}(i,j))^2,
\label{eq:eq9}
\end{equation}
where $\mathcal{L}$ is the loss function and $N$ is the total number of pixels in the image. 

By updating the parameters of generator $G$ through minimizing $\mathcal{L}$, CTA can generate  more effective cross-task perturbations.  These perturbations guide the attention of adversarial samples towards regions of high numerical value in the anti-attention maps, which are typically ignored by all visual tasks.  The detailed process of our method CTA is outlined in Algorithm 1.

\begin{table*}[htbp]
  \centering
  \caption{Performance comparison (\%) of different models for different visual tasks on clean and adversarial samples.} 
  \resizebox{\textwidth}{!}
  {
    \begin{tabular}{c|cccc|cccc|cccc}
    \toprule
    \multirow{4}[8]{*}{Attack Methods} & \multicolumn{4}{c|}{Classification Results} & \multicolumn{4}{c|}{Detection Results} & \multicolumn{4}{c}{Segmentation Results} \\
\cmidrule{2-13}          & \multicolumn{2}{c}{VGG19} & \multicolumn{2}{c|}{IncResv2} & \multicolumn{2}{c}{YOLOv3} & \multicolumn{2}{c|}{Faster-RCNN} & \multicolumn{2}{c}{DeepLabv3} & \multicolumn{2}{c}{FCN} \\
\cmidrule{2-13}          & \multicolumn{4}{c|}{Accuracy} & mAP   & mAR   & mAP   & mAR   & GCR   & mIoU  & GCR   & mIoU \\
\cmidrule{2-13}          & \multicolumn{2}{c}{$\epsilon$=10 / $\epsilon$=16} & \multicolumn{2}{c|}{$\epsilon$=10 / $\epsilon$=16} & $\epsilon$=10 / $\epsilon$=16 & $\epsilon$=10 / $\epsilon$=16 & $\epsilon$=10 / $\epsilon$=16 & $\epsilon$=10 / $\epsilon$=16 & $\epsilon$=10 / $\epsilon$=16 & $\epsilon$=10 / $\epsilon$=16 & $\epsilon$=10 / $\epsilon$=16 & $\epsilon$=10 / $\epsilon$=16 \\
    \midrule
    Clean Sample & \multicolumn{2}{c}{72.9} & \multicolumn{2}{c|}{80.8} & 59.4  & 70.9  & 51.2  & 62.8  & 94.2  & 76.3  & 93.3  & 70.3 \\
    Gaussian Noise & \multicolumn{2}{c}{70.1 / 65.7} & \multicolumn{2}{c|}{76.62 / 72.11} & 56.9 / 53.7 & 67.1 / 65.5 & 47.3 / 44.7 & 60.1 / 56.5 & 93.5 / 92.4 & 74.7 / 71.2 & 92.1 / 90.7 & 66.8 / 64.2 \\
    DR    & \multicolumn{2}{c}{67.4 / 46.17} & \multicolumn{2}{c|}{73.74 / 64.38} & 45.8 / 38 & 58.7 / 51.1 & 38.5 / 30.9 & 50.8 / 44.3 & 91 / 88.7 & 66.1 / 59.1 & 89.5 / 87.3 & 57 / 50 \\
    RB-DR & \multicolumn{2}{c}{ 65.8 / 45.11} & \multicolumn{2}{c|}{71.96 / 63.18} & 44.5 / 36.4 & 58.1 / 49.9 & 37.5 / 29.5 & 50.1 / 43.6 & 90.2 / 88.1 & 64.7 / 58.7 & 89.2 / 86.5 & 56.3 / 48.9 \\
    $\text{S}^{2}$I-FGSM & \multicolumn{2}{c}{0.8 / 0.71} & \multicolumn{2}{c|}{0.58 / 0.56} & 42.8 / 33.1 & 56.6 / 47.2 & 34.7 / 26.7 & 49 / 41.2 & 89 / 82.4 & 60 / 50.2 & 86.5 / 82 & 49.5/ 38.5 \\
    $\text{S}^{2}$I-SI-TI-FGSM & \multicolumn{2}{c}{\textbf{0.75} / \textbf{0.68}} & \multicolumn{2}{c|}{\textbf{0.54} / 0.53} & 37.7 / 27.1 & 51.5 / 40.4 & 31.4 / 18.9 & 45.6 / 32.1 & 88.3 / 80.4 & 59.2 / 38.1 & 85.9 / 80.3 & 46.4 / 29.5 \\
    CTA(ours)   & \multicolumn{2}{c}{26.52 / 7.47} & \multicolumn{2}{c|}{0.68 / \textbf{0.52}} & \textbf{31.1} / \textbf{19.5}  & \textbf{46.7} / \textbf{32.4}  & \textbf{31.1} / \textbf{16.5} & \textbf{44.9} / \textbf{28.5} & \textbf{88.1} / \textbf{77.8}  & \textbf{58.7} / \textbf{32.1} & \textbf{85.3} / \textbf{78.4} & \textbf{43.2} / \textbf{22.7} \\
    \bottomrule
    \end{tabular}%
    }
  \label{tab:normal_train}%
\end{table*}%

\begin{figure*}[t]
\centering
\includegraphics[width=1.95\columnwidth]{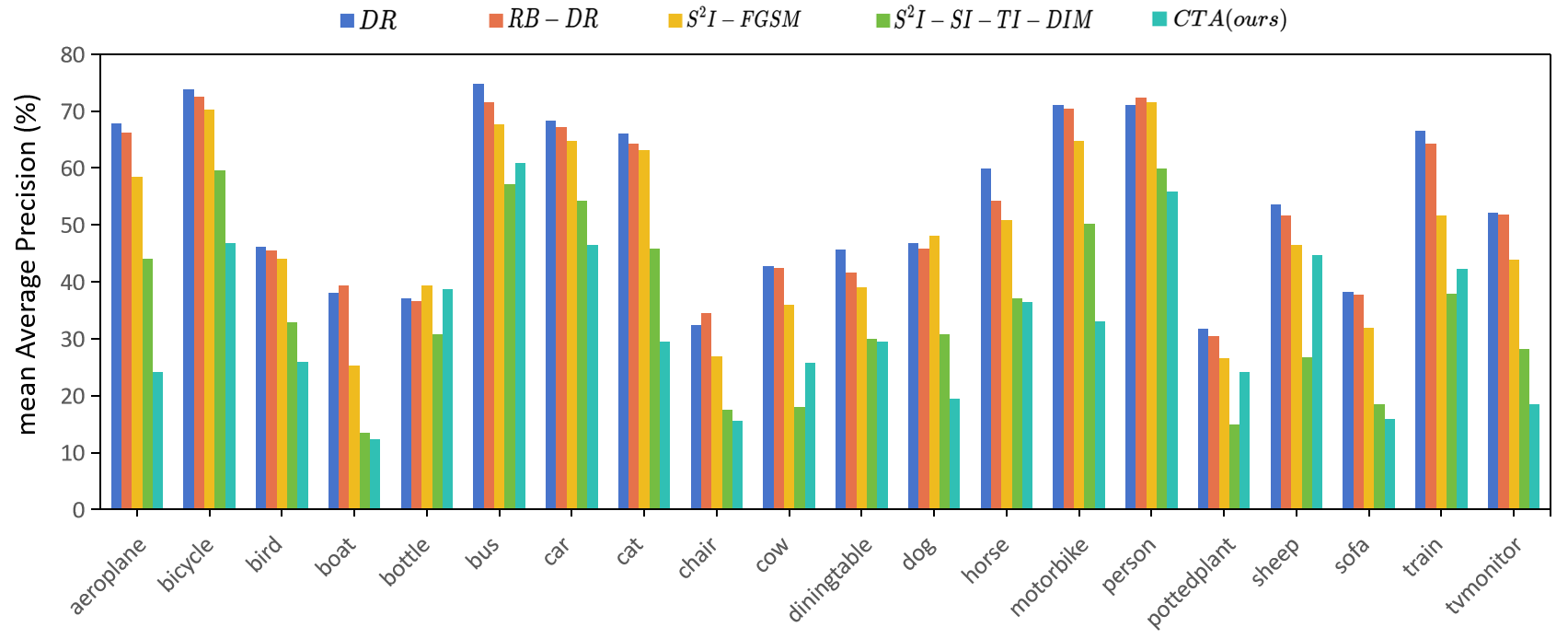}
\caption{The mAP of Faster-RCNN on the adversarial samples generated by different adversarial attack methods. The abscissa is the 20 categories of the VOC 2012 validation dataset, and the ordinate is mean Average Precision. Under the condition of  $\epsilon$ = 16, we compared the performance differences between our proposed CTA method and existing adversarial attack methods DR, RB-DR, $\text{S}^{2}$I-FGSM and $\text{S}^{2}$I-SI-TI-DIM.   }
\label{fig:det_img}
\end{figure*}

\begin{figure*}[t]
\centering
\includegraphics[width=1.95\columnwidth]{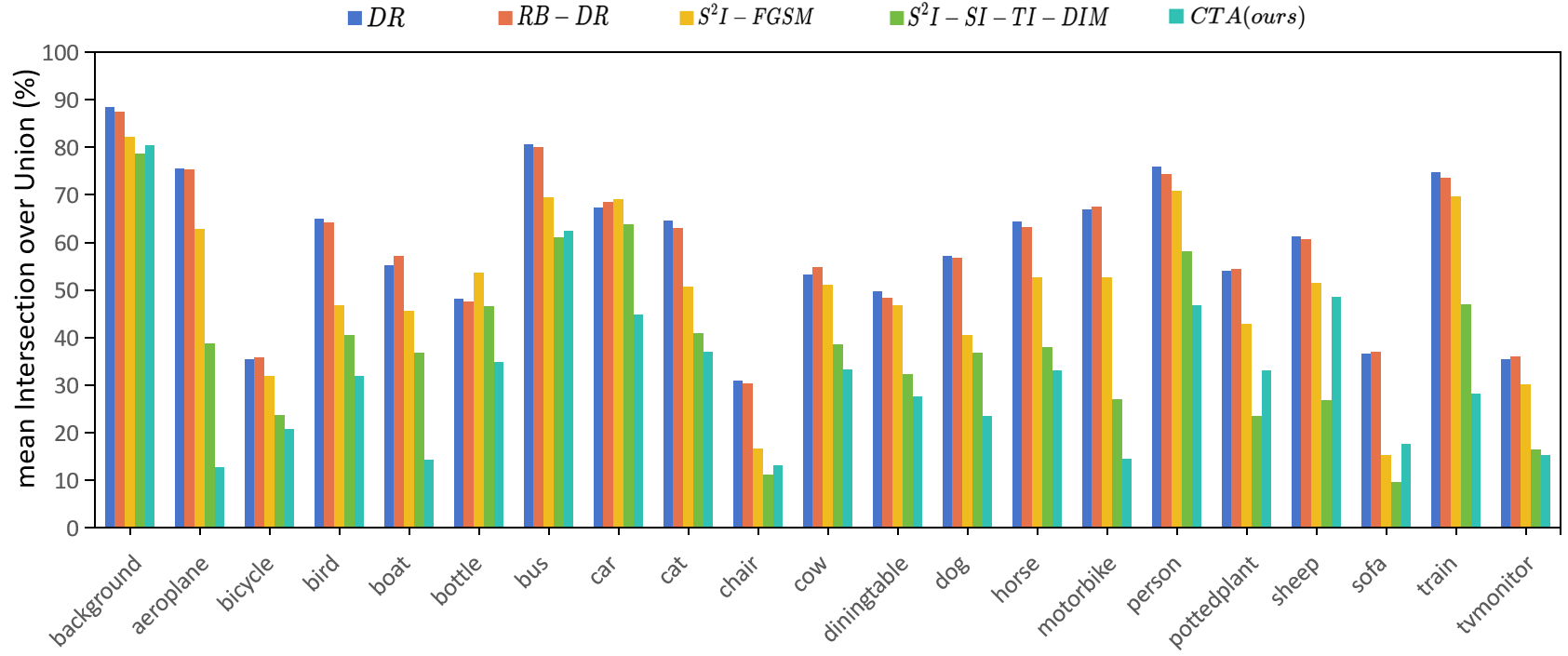}
\caption{The mIoU of DeepLabv3 on the adversarial samples generated by different adversarial attack methods. The abscissa is the 21 categories of the VOC 2012 validation dataset including the background category, and the ordinate is mean Intersection over Union. Under the condition of  $\epsilon$ = 16, we compared the performance differences between our proposed CTA method and existing adversarial attack methods DR, RB-DR, $\text{S}^{2}$I-FGSM and $\text{S}^{2}$I-SI-TI-DIM.  }
\label{fig:seg_img}
\end{figure*}

\begin{table*}[htbp]
  \centering
  \caption{Performance comparison (\%)  of different attack methods on various visual tasks in adversarial training defense models.}
    \begin{tabular}{c|cc|cc|cc}
    \toprule
    \multirow{3}[6]{*}{Attack Methods} & \multicolumn{2}{c|}{adv-IncResv2} & \multicolumn{2}{c|}{adv-Faster-RCNN} & \multicolumn{2}{c}{adv-FCN} \\
\cmidrule{2-7}          & \multicolumn{2}{c|}{Accuracy} & mAP   & mAR   & GCR   & mIoU \\
\cmidrule{2-7}          & \multicolumn{2}{c|}{$\epsilon$=10 / $\epsilon$=16} & $\epsilon$=10 / $\epsilon$=16 & $\epsilon$=10 / $\epsilon$=16 & $\epsilon$=10 / $\epsilon$=16 & $\epsilon$=10 / $\epsilon$=16 \\
    \midrule
    Clean Sample & \multicolumn{2}{c|}{80.06} & 45.9  & 58.2  & 90.6  & 59.7 \\
    Gaussian Noise & \multicolumn{2}{c|}{76.6 / 75.1} & 42.4 / 40.8 & 56.7 / 54.4 & 89.5 / 89.1 & 57.9 / 56.5 \\
    DR    & \multicolumn{2}{c|}{75.85 / 69.05} & 36.8 / 34 & 50 / 52.6 & 89.9 / 88.5 & 58.7 / 52.1 \\
    RB-DR & \multicolumn{2}{c|}{74.66 / 67.64} & 35.7 / 32.4 & 49.2  / 47.9 & 89.4 / 87.9 & 57.8 / 50.4 \\
    $\text{S}^{2}$I-FGSM & \multicolumn{2}{c|}{0.97 / 1.01} & 34.1 / 27.9 & 48.1 / 41.6 & 88 / 86 & 51.4 / 45.5 \\
    $\text{S}^{2}$I-SI-TI-FGSM & \multicolumn{2}{c|}{\textbf{0.88} / \textbf{0.91}} & 33.6 / 25.5 & 47.7 / 37.8 & 87.4 / 85.4 & 50.5 / 43.2 \\
    CTA   & \multicolumn{2}{c|}{1.04 / 0.93} & \textbf{32.6} / \textbf{19.4} & \textbf{46.3} / \textbf{31} & \textbf{85.9} / \textbf{79.3} & \textbf{46.7} / \textbf{24.1} \\
    \bottomrule
    \end{tabular}%
  \label{tab:adv}%
\end{table*}%

\section{Experiments}

In this section, we conduct extensive quantitative experiments on three classic visual tasks: image classification, object detection, and semantic segmentation, to evaluate the effectiveness and robustness of our proposed method CTA for cross-task attack. We also perform qualitative experiments to observe the trend of Grad-CAM attention heatmaps across the training iterations.

\subsection{Experimental Setup}

\subsubsection{Evaluation datasets.}

Following the experimental settings of previous work \cite{Dong2018, Lu2020, Zhang2021},  we randomly select 10 samples from 1000 classes in the ImageNet validation set, totaling 10000 samples, as the validation set for the classification task. We use the complete validation set of  PASCAL VOC 2012 as the validation set for object detection and semantic segmentation tasks.

\subsubsection{Generator training.}
We use a ResNet architecture composed of downsampling blocks and upsampling blocks as generator $G$ \cite{Johnson2016}. We use the images in the VOC 2012 training set to train the generator $G$. It is worth noting that the training of CTA does not require any ground true labels, as we use ready-made models to extract anti-attention graphs for self-supervised training. The pretrained model for each visual task adopts classic and ready-made models, with ResNet50 as classification model $f_{cls}$, SSD as detection model $f_{det}$, and U-nets as segmentation model $f_{s}$. The  feature extractor $D$ for adversarial samples adopts ResNet50. We use Adam optimizer for training, learning rate is set to 1e-3, first and second moment exponential decay rates are set to 0.5 and 0.99. We train two versions of perturbation generator $G$ based on different perturbation range thresholds, corresponding to epsilon 10 and 16 respectively. Our experimental device uses three GPU of RTX2080ti with 11GB memory and a CPU of Intel(R) Core(TM) i5-10400F.

\subsubsection{Comparison attack algorithms.}
We choose five adversarial attack methods for comparison:  1. DR, a cross-task adversarial attack method that does not rely on any specific task loss function, it reduces the feature map standard deviation to create adversarial examples that fool multiple visual tasks; 2. RB-DR, which adds a random blur (RB) data augmentation method on the basis of DR to increase the success rate of attack. 3. $\text{S}^{2}$I-FGSM, a single-task cross-model attack algorithm that belongs to the FGSM adversarial attack family, it relies on a specific task loss function, and uses frequency domain transformation to enhance the transferability of adversarial examples and improve the cross-model attack effect. To the best of our knowledge, $\text{S}^{2}$I-FGSM has been proven to be the state-of-the-art method for single-task cross-model attack. 4. $\text{S}^{2}$I-SI-TI-DIM, where we have aggregated existing popular single-task  attack methods, including $\text{S}^{2}$I-FGSM, TI-FGSM, SI-FGSM and DIM, to achieve the strongest single-task cross-model attack for comparison.  5. Gaussian noise, the performance baseline for adversarial attacks. The hyperparameter settings for $\text{S}^{2}$I-FGSM and $\text{S}^{2}$I-SI-TI-DIM are set according to the default settings in $\text{S}^{2}$I-FGSM \cite{Long2022}. The hyperparameter settings for DR and RB-DR are set according to the default settings in DR \cite{Lu2020}.

\subsubsection{Evaluation metric.}
For the classification task of Imagenet, we use Top-1 accuracy as the evaluation metric. For the obeject detection task of PASCAL VOC 2012, we use mean Average Precision (mAP) and mean Average Recall (mAR) as evaluation metrics. For the semantic segmentation task of PASCAL VOC 2012, we use Global Correct Rate (GCR) and mean Intersection over Union (mIoU) as evaluation metrics.

\subsection{Attack Normally Trained Models}
\subsubsection{Image classification task results.}
In the image classification task, we choose VGG19 \cite{simonyan2014very} and IncResv2 \cite{szegedy2017inception}  pretrained on ImageNet as the attack target models. Table \ref{tab:normal_train} shows the classification accuracy of our proposed CTA attack method and the compared attack methods on the ImageNet validation set. It can be observed that the cross-task attack method DR performs very weakly in classification tasks, only reducing the accuracy rate by average 13.92\% compared to clean samples. After applying random blur (RB) on the basis of DR, RB-DR has an about 2\% improvement in attack performance. $\text{S}^{2}$I-FGSM and $\text{S}^{2}$I-SI-TI-DIM are adversarial attack methods designed for classification, which use the loss function and real labels of the classification task, thus having very strong attack performance in classification tasks. We regard  $\text{S}^{2}$I-FGSM and $\text{S}^{2}$I-SI-TI-DIM as the upper bound  of attack performance for classification tasks. Compared to DR, our CTA reduces the accuracy rate by 54.08\% and is close to the upper bound  of classification attack performance, demonstrating its effectiveness in classification scenarios.

\begin{figure*}[t]
\centering
\includegraphics[width=1.7\columnwidth]{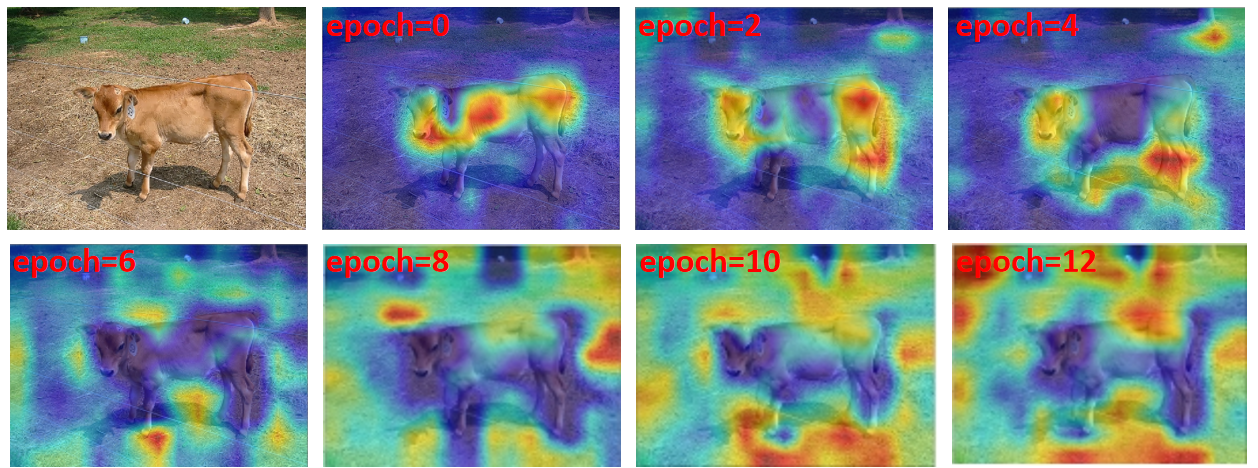}
\caption{Grad-CAM heatmap of adversarial samples with different training iterations in CTA method.}
\label{fig:gradcam}
\end{figure*}

\subsubsection{Object detection task results.}
In the object detection task, we choose YOLOv3 \cite{redmon2018yolov3} and Faster-RCNN \cite{ren2015faster} pretrained on PASCAL VOC 2012 as the attack target models. Table \ref{tab:normal_train} shows the mAP and mAR of our proposed CTA attack method and the compared attack methods on PASCAL VOC 2012 validation set. Figure \ref{fig:det_img} shows the mAP of 20 categories of Faster-RCNN on different adversarial samples.  As shown in Table \ref{tab:normal_train}, DR can reduce the mAP and mAR by an average of 20.6\% and 15.6\% compared to clean image.  Random blur (RB-DR)  can slightly improve DR's attack performance.  The attack performance of $\text{S}^{2}$I-FGSM is similar to RB-DR, but the performance of $\text{S}^{2}$I-SI-TI-DIM is significantly better  compared to $\text{S}^{2}$I-FGSM. The reason is that $\text{S}^{2}$I-FGSM, TI-FGSM, SI-FGSM and DIM are designed to improve transfer ability, so their combined method  $\text{S}^{2}$I-SI-TI-DIM has stronger transfer ability than any single component, making it perform well in cross-task scenarios.  Compared to existing attack methods, our CTA achieves the lowest mAP and mAR in all cases. As shown in Figure  \ref{fig:det_img}, CTA has the lowest mAP in 14 out of 20 categories. Our experiments  demonstrate CTA's effectiveness in object detection scenarios.

\subsubsection{Semantic segmentation task results.}
In semantic segmentation, we choose DeepLabv3 \cite{chen2018encoder} and FCN  \cite{long2015fully} pretrained on PASCAL VOC 2012 as the attack target models. Table \ref{tab:normal_train} shows the GCR and mIoU of our proposed CTA attack method and the compared attack methods on PASCAL VOC 2012 validation set. Figure \ref{fig:seg_img} show the mIoU of 21 categories of deeplabv3 on different adversarial samples. As shown in Table \ref{tab:normal_train}, DR can reduce the GCR and mIoU  by an average of 4.62\% and 15.25\% compared to clean image. Random blur (RB-DR)  can slightly improve DR's attack performance. $\text{S}^{2}$I-FGSM and $\text{S}^{2}$I-SI-TI-DIM exhibit greater attack performance than DR and RB-DR due to their strong transfer ability. Compared to existing attack methods, our CTA achieves the lowest GCR and mIoU in all cases. As shown in Figure \ref{fig:seg_img}, CTA has the lowest mIoU in 15 out of 21 categories. Our experiments demonstrate CTA's effectiveness in semantic segmentation scenarios.

\subsection{Attack Adversarially Trained Defense Models}
A deep learning model trained on adversarial examples can weaken the effectiveness of adversarial attacks. To further demonstrate the effectiveness of our adversarial attack method, we conducted experiments on attacking defense models in classification, detection, and segmentation tasks. Referring to the work \cite{tramer2018ensemble}, we conduct ensemble adversarial training on IncResv2, Faster-RCNN and FCN to obtained dense models adv-incResv2, adv-Faster-RCNN and adv-FCN for different tasks. As shown in Table \ref{tab:adv}, it can be seen that DR and RB-DR have lost their ability to attack defense models for classification and segmentation tasks under the condition of $\epsilon$=10. Compared with existing attack methods, our proposed CTA method is still the best in object detection and semantic segmentation scenarios, and is also very close to $\text{S}^{2}$I-FGSM and $\text{S}^{2}$I-SI-TI-DIM in classification tasks.

\subsection{Attention Visualization}
We visualized the Grad-CAM heatmap of the adversarial samples corresponding to each epoch of the training iteration. As shown in Figure \ref{fig:gradcam}, as the number of training iterations increases, the attention of non-important regions (i. e., backgrounds) that should be ignored becomes high, while the attention of important regions (i. e., cow) that should be paid attention to becomes low. This experiment intuitively demonstrates the attention movement process of adversarial samples using our CTA.

\section{Conclusion}
In this paper, we propose a novel cross-task adversarial attack method CTA, which can generate adversarial examples that can fool multiple visual tasks simultaneously. Unlike existing attack methods, CTA can directionally guide the attention shift of adversarial samples. CTA utilizes Grad-CAM to extract common attention regions from different visual task models, and uses a generator to generate adversarial samples that can shift attention to areas overlooked by all visual tasks, thereby achieving cross-task attacks. CTA does not rely on any specific task loss function or ground true label, making it a general and flexible method for cross-task attack. Our extensive experiments have shown that our method outperforms the comparative methods in object detection and semantic segmentation tasks. In image classification task, our method outperforms existing cross-task attack methods and approaches the single-task attack methods designed for classification task. We also visualize the Grad-CAM attention heatmaps of our method CTA, and intuitively demonstrates the attention movement process of adversarial samples with increasing training iterations.

\section*{Acknowledgments}
Supported in part by the National Natural Science Foundation of China under Grant 62276222.


\bibliographystyle{IEEEtran}
\bibliography{IEEEabrv, reference}

\vspace{12pt}

\end{document}